\documentclass[]{article}
\usepackage{proceed2e}
\usepackage[protrusion=true,spacing=true]{microtype}
\usepackage{amsmath,amssymb,amsfonts,amsthm}
\usepackage{algorithmic,algorithm}
\usepackage{graphicx,tikz}\usetikzlibrary{positioning}
\usepackage{natbib,cite}
\usepackage{enumitem}
\usepackage{times}

\usepackage[absolute,overlay]{textpos}
\newtheoremstyle{mystyle}
  {5pt}		
  {0pt}		
  {}			
  {}			
  {\bfseries}	
  {.}			
  { }			
  {}			
\theoremstyle{mystyle}

\newtheorem{_definition}{Definition}
\newtheorem{_assumption}{Assumption}
\newtheorem{_theorem}{Theorem}
\newtheorem{_example}{Example}
\newtheorem{_property}{Property}

\newcommand{\td}{\Delta}
\newcommand{\pr}{\text{Pr}}
\newcommand{\ao}{\text{AO}}
\newcommand{\as}{\text{AS}}
\newcommand{\leftm}{14pt}
\newcommand{\state}{\STATE\hspace{-5pt}}
\newcommand{\hist}{H^\tau\hspace{-2pt},\hspace{-1pt}H^\infty\hspace{-1pt}}

\title{On Convergence and Optimality of Best-Response Learning \\ with Policy Types in Multiagent Systems}

\author{
	\textbf{Stefano V. Albrecht}\hspace{1pt} \\
	School of Informatics \\
	University of Edinburgh \\
	Edinburgh EH8 9AB, UK \\
	\texttt{s.v.albrecht@sms.ed.ac.uk}
	\And
	\hspace{1pt}\textbf{Subramanian Ramamoorthy} \\
	School of Informatics \\
	University of Edinburgh \\
	Edinburgh EH8 9AB, UK \\
	\texttt{s.ramamoorthy@ed.ac.uk}
}

\begin{document}

	\maketitle

	\begin{textblock}{49}(7.5,2)
		\small This is a corrected version of the original paper published in the \textit{Proceedings of the 30th Conference on Uncertainty in Artificial Intelligence}. The changes pertain to Section~\ref{sec:product}, specifically Theorem~\ref{theo:p-post}. See the appendix for more details \citep{ar2014app}.
	\end{textblock}

	\begin{abstract}
While many multiagent algorithms are designed for homogeneous systems (i.e. all agents are identical), there are important applications which require an agent to coordinate its actions without knowing a priori how the other agents behave. One method to make this problem feasible is to assume that the other agents draw their latent policy (or type) from a specific set, and that a domain expert could provide a specification of this set, albeit only a partially correct one. Algorithms have been proposed by several researchers to compute posterior beliefs over such policy libraries, which can then be used to determine optimal actions. In this paper, we provide theoretical guidance on two central design parameters of this method: Firstly, it is important that the user choose a posterior which can learn the true distribution of latent types, as otherwise suboptimal actions may be chosen. We analyse convergence properties of two existing posterior formulations and propose a new posterior which can learn correlated distributions. Secondly, since the types are provided by an expert, they may be inaccurate in the sense that they do not predict the agents' observed actions. We provide a novel characterisation of optimality which allows experts to use efficient model checking algorithms to verify optimality of types.
	\end{abstract}

	\section{INTRODUCTION} \label{sec:intro}

Many multiagent algorithms are developed with a homogeneous setting in mind, meaning that all agents use the same algorithm and are a priori aware of this fact. However, there are important applications for which this assumption may not be adequate, such as human-machine interaction, robot search and rescue, and financial markets. In such problems, it is important that an agent be able to effectively coordinate its actions without knowing \emph{a priori} how the other agents behave. The importance of this problem has been discussed in works such as \citep{ar2013,skkr2010,bm2005}.

This problem is hard since the agents may exhibit a large variety of behaviours. General-purpose algorithms for multiagent learning are often impracticable, either because they take too long to produce effective policies or because they rely on prior coordination of behaviours \citep{ar2012}. However, it has been recognised (e.g. \citep{ar2013,bsk2011}) that the complexity of this problem can often be reduced by assuming that there is a latent set of policies for each agent and a latent distribution over these policies, and that a domain expert can provide informed guesses as to what the policies might be. (These guesses could also be generated automatically, e.g. using some machine learning method on a corpus of historical data.)

One algorithm that takes this approach is {\textit{Harsanyi-Bellman Ad Hoc Coordination}} (HBA) \citep{ar2013}. This algorithm maintains a set of user-defined types (by ``type'', we mean a policy or programme which specifies the behaviour of an agent) over which it computes posterior beliefs based on the agents' observed actions. The beliefs are then used in a planning procedure to compute expected payoffs for all actions (a procedure combining the concepts of Bayesian Nash equilibrium and Bellman optimality) and the best action is chosen. HBA was implemented as a reinforcement learning procedure and shown to be effective in both simulated and human-machine problems \citep{ar2013}. Similar algorithms were studied in \citep{bsk2011,cm1999}.

While works such as \citep{ar2013,bsk2011,cm1999} demonstrate the practical usefulness of such methods, they provide no theoretical guidance on two central design parameters: Firstly, one may compute the posterior beliefs in various ways, and it is important that the user choose a posterior formulation which is able to accurately approximate the latent distribution of types. This is important as otherwise the expected payoffs may be inaccurate, in which case HBA may choose suboptimal actions. In this paper, we analyse the convergence conditions of two existing posterior formulations and we propose a new posterior which can learn correlated type distributions. These theoretical insights can be applied by the user to choose appropriate posteriors.

Secondly, since the types are provided by the user (or generated automatically), they may be inaccurate in the sense that their predictions deviate from the agents' observed actions. This raises the need for a theoretical analysis of how much and what kind of inaccuracy is acceptable for HBA to be able to solve its task, by which we mean that it drives the system into a terminal state. (A different question pertains to payoff maximisation; we focus on task accomplishment as it already includes many practical problems.) We describe a methodology in which we formulate a series of desirable termination guarantees and analyse the conditions under which they are met. Furthermore, we provide a novel characterisation of optimality which is based on the notion of probabilistic bisimulation \citep{ls1991}. In addition to concisely defining what constitutes optimal type spaces, this allows the user to apply efficient model checking algorithms to verify optimality in practice.

	\section{RELATED WORK} \label{sec:relwork}

Opponent modelling methods such as case-based reasoning \citep{gs2001} and recursive modelling \citep{gd2000} are relevant to the extent that they can complement the user-defined types by creating new types (the opponent models) on the fly. For example, \citep{ar2013} used a variant of case-based reasoning and \citep{bsk2011} used a tree-based classifier to complement the user-defined types.

Plays and play books \citep{bm2005} are similar in spirit to types and type spaces. However, plays specify the behaviour of an entire team, with additional structure such as applicability and termination conditions, and roles for each agent. In contrast, types specify the action probabilities of a single agent and do not require commitment to conditions and roles.

Plans and plan libraries \citep{c2001} are conceptually similar to types and type spaces. However, the focus of plan recognition has been on identifying the goal of an agent (e.g. \citep{bkd2009}) and efficient representation of plans (e.g. \citep{ak2007}), while types are used primarily to compute expected payoffs and can be efficiently represented as programmes \citep{ar2013,bsk2011}.

I-POMDPs \citep{gd2005} and I-DIDs \citep{dzc2009} are related to our work since they too assume that agents have a latent type. These methods are designed to handle the full generality of partially observable states and latent types, and they explicitly model nested beliefs. However, this generality comes at a high computational cost and the solutions are infeasible to compute in many cases. In contrast, we remain in the setting of fully observable states, and we implicitly allow for complex beliefs within the specification of types. This allows our methods to be computationally more tractable.

To the best of our knowledge, none of these related works directly address the theoretical questions considered in this paper. While our results apply to \citep{ar2013,bsk2011,cm1999}, we believe they could be generalised to account for some of the other related works as well. This includes the methodology described in Section~\ref{sec:type-space}.

	\section{PRELIMINARIES} \label{sec:prel}

		\subsection{MODEL}

Our analysis is based on the stochastic Bayesian game \citep{ar2013}:

\begin{_definition}
	A \emph{stochastic Bayesian game} (SBG) consists~of
	\vspace{-20pt}
	\begin{itemize}[leftmargin=\leftm,itemsep=0pt]
		\item discrete state space $S$ with initial state $s^0 \in S$ and \newline terminal states $\bar{S} \subset S$
		\item players $N = \left\{ 1,...,n \right\}$ and for each $i \in N$:
		\vspace{-4pt}
		\begin{itemize}[leftmargin=\leftm,itemsep=1pt]
			\item set of actions $A_i$ (where $A = A_1 \times ... \times A_n$)
			\item type space $\Theta_i$ (where $\Theta = \Theta_1 \times ... \times \Theta_n$)
			\item payoff function $u_i : S \times A \times \Theta_i \rightarrow \mathbb{R}$
			\item strategy $\pi_i : \mathbb{H} \times A_i \times \Theta_i \rightarrow [0,1]$
		\end{itemize}
		\item state transition function $T : S \times A \times S \rightarrow [0,1]$
		\item type distribution $\td : \Theta \rightarrow [0,1]$
	\end{itemize}
	\vspace{-6pt}
	where $\mathbb{H}$ contains all \emph{histories} $H^t = \langle s^0,a^0,s^1,a^1,...,s^t \rangle$ with $t \geq 0$, $(s^\tau,a^\tau) \in S \times A$ for $0 \leq \tau < t$, and $s^t \in S$.
\end{_definition}

\vspace{3pt}

\begin{_definition}
	A SBG starts at time $t = 0$ in state $s^0$:
	\vspace{-7pt}
	\begin{enumerate}[leftmargin=\leftm,itemsep=0pt]
		\item In state $s^t$, the types $\theta_1^t,...,\theta_n^t$ are sampled from $\Theta$ with probability $\td (\theta_1^t,...,\theta_n^t)$, and each player $i$ is informed only about its own type $\theta_i^t$.
		\item Based on the history $H^t$, each player $i$ chooses an action $a_i^t \in A_i$ with probability $\pi_i(H^t,a_i^t,\theta_i^t)$, resulting in the joint action $a^t = (a^t_1,...,a^t_n)$.
		\item The game transitions into a successor state $s^{t+1}\hspace{-1pt}\in~\hspace{-2pt}S$ with probability $T(s^t,a^t,s^{t+1})$, and each player $i$ receives an individual payoff given by $u_i(s^t,a^t,\theta_i^t)$.
	\end{enumerate}
	\vspace{-6pt}
	This process is repeated until a terminal state $s^t \in \bar{S}$ is reached, after which the game stops.
\end{_definition}

		\subsection{ASSUMPTIONS}

We make the following general assumptions in our analysis:

\begin{_assumption}
	We control player $i$, by which we mean that we choose the strategies $\pi_i$ (using HBA). Hence, player $i$ has only one type, $\theta_i$, which is known to us.
\end{_assumption}

We sometimes omit $\theta_i$ in $u_i$ and $\pi_i$ for brevity, and we use $j$ and $-i$ to refer to the other players (e.g. $A_{-i} = \times_{j \neq i} \, A_j$).

\begin{_assumption}
	Given a SBG $\Gamma$, we assume that all elements of $\Gamma$ are known except for the type spaces $\Theta_j$ and the type distribution $\td$, which are \emph{latent variables}.
\end{_assumption}

\begin{_assumption}
	We assume \emph{full observability} of states and actions. That is, we are always informed of the current history $H^t$ before making a decision.
\end{_assumption}

\begin{_assumption} \label{ass:random}
	For any type $\theta_j$ and history $H^t$, there exists a \emph{unique} sequence $(\chi_{a_j})_{a_j \in A_j}$ such that $\pi_j(H^t,a_j,\theta_j) = \chi_{a_j}$ for all $a_j \in A_j$.
\end{_assumption}

We refer to this as \emph{external} randomisation and to the opposite (when there is no unique $\chi_{a_j}$) as \emph{internal} randomisation. Technically, Assumption~\ref{ass:random} is implied by the fact that $\pi_j$ is a function, which means that any input is mapped to exactly one output. However, in practice this can be violated if randomisation is used ``inside'' a type implementation, hence it is worth stating it explicitly. Nonetheless, it can be shown that under full observability, external randomisation is equivalent to internal randomisation. Hence, Assumption~\ref{ass:random} does not limit the types we can represent.

\begin{_example}
	Let there be two actions, A and B, and let the expected payoffs for agent $i$ be $E(A) > E(B)$. The agent uses $\epsilon$-greedy action selection \citep{sb1998} with $\epsilon > 0$. If agent $i$ randomises \emph{externally}, then the strategy $\pi_i$ will assign action probabilities $\langle 1-\epsilon/2,\epsilon/2 \rangle$. If the agent randomises \emph{internally}, then with probability $\epsilon$ it will assign probabilities $\langle 0.5,0.5 \rangle$ and with probability $1-\epsilon$ it will assign $\langle 1,0 \rangle$, which is equivalent to external randomisation.
\end{_example}

		\subsection{ALGORITHM}

Algorithm~\ref{alg:hba} gives a formal definition of HBA (based on \citep{ar2013}) which is the central algorithm in this analysis. (Section~\ref{sec:intro} provides an informal description.) Throughout this paper, we will use $\Theta_j^*$ and $\pr_j$, respectively, to denote the user-defined type space and posterior for player~$j$, where $\pr_j (\theta_j^* | H^t)$ is the probability that player $j$ has type $\theta_j^* \in \Theta_j^*$ after history $H^t$. Furthermore, we will use $\pr$ to denote the \emph{combined} posterior, with $\pr (\theta_{-i}^* | H^t) = \prod_{j \neq i} \pr_j (\theta_j^* | H^t)$, and we sometimes refer to this simply as \emph{the posterior}.

Note that the likelihood $L$ in \eqref{eq:post} is unspecified at this point. We will consider two variants for $L$ in Section~\ref{sec:td}. The prior probabilities $P_j(\theta_j^*)$ in \eqref{eq:post} can be used to specify prior beliefs about the distribution of types. It is convenient to specify $\pr_j (\theta_j^* | H^t) = P_j(\theta_j^*)$ for $t = 0$. Finally, note that \eqref{eq:bne}/\eqref{eq:bellman} define an infinite regress. In practice, this may be implemented using stochastic sampling (e.g. as in \citep{ar2013,bsk2011}) or by terminating the regress after some finite amount of time. In this analysis, we assume that \eqref{eq:bne}/\eqref{eq:bellman} are implemented as given.

\begin{algorithm}[t]
	\small
	\begin{algorithmic}
		\state \textbf{Input:} \hspace{7pt} SBG $\Gamma$, player $i$, user-defined type spaces $\Theta_j^*$,
		\state \hspace{34pt} history $H^t$, discount factor $0 \leq \gamma \leq 1$ \\[5pt]
		\state \textbf{Output:} \ Action probabilities $\pi_i(H^t,a_i)$ \\[5pt]
		\state \textbf{\textit{1.}} For each $j \neq i$ and $\theta_j^* \in \Theta_j^*$, compute posterior probability
			\begin{equation} \label{eq:post}
				\pr_j (\theta_j^* | H^t) = \frac{L(H^t | \theta_j^*) \, P_j(\theta_j^*)}{\sum_{\hat{\theta}_j^* \in \Theta_j^*} L(H^t | \hat{\theta}_j^*) \, P_j(\hat{\theta}_j^*)}
			\end{equation}
		\state \textbf{\textit{2.}} For each $a_i \in A_i$, compute expected payoff $E_{s^t}^{a_i}(H^t)$ with
			\begin{equation} \label{eq:bne}
				\hspace{-5pt} E_s^{a_i}(\hat{H}) = \hspace{-19pt} \sum_{\hspace{18pt} \theta^*_{-i} \in \Theta^*_{-i}} \hspace{-18pt} \pr (\theta^*_{-i} | H^t) \hspace{-17pt} \sum_{\hspace{18pt} a_{-i} \in A_{-i}} \hspace{-19pt} Q_s^{a_{i,-i}}(\hat{H}) \, \prod_{j \neq i} \hspace{-1pt} \pi_j(\hat{H},a_j,\theta^*_j)
			\end{equation}\vspace{1pt}
			\begin{equation} \label{eq:bellman}
				\hspace{-5pt} Q_s^a(\hat{H}) = \hspace{-2pt} \sum_{s' \in S} T(s,a,s') \left[ u_i(s,a) + \gamma \max_{a_i}E_{s'}^{a_i} \hspace{-2pt} \left( \langle \hat{H},a,s' \rangle \right) \right]
			\end{equation}\vspace{1pt}
		\state where $\pr (\theta_{-i}^* | H^t) = \prod_{j \neq i} \pr_j (\theta_j^* | H^t)$ and $a_{i,-i} \triangleq (a_i,a_{-i})$ \\[5pt]
		\state \textbf{\textit{3.}} Distribute $\pi_i(H^t,\cdot)$ uniformly over $\arg\max_{a_i} E_{s^t}^{a_i}(H^t)$
	\end{algorithmic}
	\caption{\small Harsanyi-Bellman Ad Hoc Coordination (HBA) \citep{ar2013}}
	\label{alg:hba}
\end{algorithm}

	\section{LEARNING THE TYPE DISTRIBUTION} \label{sec:td}

This section is concerned with convergence and correctness properties of the posterior. The theorems in this section tell us if and under what conditions HBA will learn the type distribution of the game. As can be seen in Algorithm~\ref{alg:hba}, this is important since the accuracy of the expected payoffs \eqref{eq:bne} depends crucially on the accuracy of the posterior \eqref{eq:post}.

However, for this to be a well-posed learning problem, we have to assume that the posterior $\pr$ can refer to the same elements as the type distribution $\td$. Therefore, the results in this section pertain to a weaker form of \emph{ad hoc coordination} \citep{ar2013} in which the user knows that the latent type space $\Theta_j$ must be a subset of the user-defined type space $\Theta_j^*$. Formally, we assume:

\begin{_assumption} \label{as:inclusion}
	$\forall j \neq i : \Theta_j \subseteq \Theta_j^*$
\end{_assumption}

Based on this assumption, we simplify the notation in this section by dropping the * in $\theta_j^*$ and $\Theta_j^*$. The general case in which Assumption~\ref{as:inclusion} does \emph{not} hold is addressed in Section~\ref{sec:type-space}.

We consider two kinds of type distributions:

\begin{_definition}
	A type distribution $\td$ is called \emph{pure} if there is $\theta \in \Theta$ such that $\td(\theta) = 1$. A type distribution is called \emph{mixed} if it is not pure.
\end{_definition}

Pure type distributions can be used to model the fact that each player has a fixed type throughout the game, e.g. as in \citep{bsk2011}. Mixed type distributions, on the other hand, can be used to model randomly changing types. This was shown in \citep{ar2013}, where a mixed type distribution was used to model defective agents and human behaviour.

		\subsection{PRODUCT POSTERIOR} \label{sec:product}

We first consider the product posterior:

\begin{_definition}
	The \emph{product posterior} is defined as \eqref{eq:post} with
	\begin{equation} \label{eq:p-post}
		L(H^t | \theta_j) = \prod_{\tau = 0}^{t-1} \pi_j(H^\tau,a_j^\tau,\theta_j)
	\end{equation}
\end{_definition}

This is the standard posterior formulation used in Bayesian games (e.g. \citep{kl1993}) and was used in \citep{ar2013,bsk2011}.

It can be shown that, under a pure type distribution and if HBA does not a priori rule out any of the types in $\Theta_j^*$, then it will learn to make correct future predictions. Let $H^\infty$ be an infinite history with prefix $H^\tau$, and denote by $P_\td(\hist)$ and $P_\pr(\hist)$, respectively, the \emph{true} probability (based on $\td$) and the probability assigned by HBA (based on $\pr$) that $H^\tau$ will continue as prescribed by $H^\infty$.

\begin{_theorem} \label{theo:p-post}
	Let $\Gamma$ be a SBG with a pure type distribution $\td$. If HBA uses a product posterior and if the prior probabilities $P_j$ are positive ($\forall \theta_j^* \in \Theta_j^* : P_j(\theta_j^*) > 0$), then: \newline for any $\epsilon > 0$, there is a time $t$ from which ($\tau \geq t$) \newline $P_\pr(\hist)(1\hspace{-2pt}-\hspace{-2pt}\epsilon) \leq P_\td(\hist) \leq (1\hspace{-2pt}+\hspace{-2pt}\epsilon)P_\pr(\hist)$ for all $H^\infty$ with $P_\td(\hist) > 0$.
\end{_theorem}

\begin{proof}
	The proof extends the convergence result of \citep{kl1993}. A full proof is provided in the appendix document \citep{ar2014app}.
\end{proof}

Unfortunately, there is a subtle but important asymmetry between making correct future predictions and knowing the true type distribution: while the latter implies the former, examples can be created to show that the reverse is not true in general. Therefore, while HBA is guaranteed to make correct future predictions after some time, it is not guaranteed to learn the type distribution of the game.

Note that Theorem~\ref{theo:p-post} pertains to pure type distributions only. The following example shows that the product posterior may fail in SBGs with mixed type distributions:

\begin{_example} \label{ex:product-mixed}
	Consider a SBG with two players. Player 1 is controlled by HBA using a product posterior while player 2 has two types, $\theta_A$ and $\theta_B$, which are assigned by a mixed type distribution $\td$ with $\td(\theta_A) = \td(\theta_B) = 0.5$. The type $\theta_A$ always chooses action A while $\theta_B$ always chooses action B. In this case, there will be a time $t$ after which both types have been assigned at least once, and so both actions A and B have been played at least once by player 2. This means that from time $t$ and all subsequent times $\tau \geq t$, we have $\pr_2 (\theta_A | H^\tau) = \pr_2 (\theta_B | H^\tau) = 0$ (that is, $\pr_2$ is undefined), and HBA will fail to make correct future predictions.
\end{_example}

		\subsection{SUM POSTERIOR}

We now consider the sum posterior:

\begin{_definition}
	The \emph{sum posterior} is defined as \eqref{eq:post} with
	\begin{equation} \label{eq:s-post}
		L(H^t | \theta_j) = \sum_{\tau = 0}^{t-1} \pi_j(H^\tau,a_j^\tau,\theta_j)
	\end{equation}
\end{_definition}

The sum posterior was introduced in \citep{ar2013} to allow HBA to recognise changed types. In other words, the purpose of the sum posterior is to learn mixed type distributions. It is easy to see that a sum posterior would indeed learn the mixed type distribution in Example~\ref{ex:product-mixed}. However, we now give an example to show that, without additional requirements, the sum posterior does not necessarily learn any (pure or mixed) type distribution:

\begin{_example} \label{ex:sum-pure}
	Consider a SBG with two players. Player 1 is controlled by HBA using a sum posterior while player 2 has two types, $\theta_A$ and $\theta_{AB}$, which are assigned by a pure type distribution $\td$ with $\td(\theta_A) = 1$. The type $\theta_A$ always chooses action A while $\theta_{AB}$ chooses actions A and B with equal probability. While the product posterior converges to the correct probabilities $\td$, the sum posterior converges to probabilities $\langle \frac{2}{3} , \frac{1}{3} \rangle$, which is incorrect.
\end{_example}

Note that this example can be readily modified to use a mixed type distribution, with similar results. Therefore, we conclude that, in general, the sum posterior does not necessarily learn any type distribution.

Under what condition is the sum posterior guaranteed to learn the true type distribution? Consider the following two quantities, which can be computed from a given history $H^t$:

\begin{_definition}
	The \emph{average overlap} of player $j$ in $H^t$ is
	\begin{equation} \label{eq:ao}
		\ao_j(H^t) = \frac{1}{t} \sum_{\tau = 0}^{t-1} \left[ | \Lambda_j^\tau | \geq 2 \right] _1 \sum_{\theta_j \in \Theta_j} \hspace{-2pt} \pi_j(H^\tau,a_j^\tau,\theta_j) \, |\Theta_j|^{-1}
	\end{equation}
	\begin{equation*}
		\Lambda_j^\tau = \left\{ \theta_j \in \Theta_j \, | \, \pi_j(H^\tau,a_j^\tau,\theta_j) > 0 \right\} \vspace{4pt}
	\end{equation*}
	where $[b]_1 = 1$ if $b$ is true, else $0$.
\end{_definition}

\begin{_definition}
	The \emph{average stochasticity} of player $j$ in $H^t$ is
	\begin{equation} \label{eq:as}
		\as_j(H^t) = \frac{1}{t} \sum_{\tau = 0}^{t-1} |\Theta_j|^{-1} \hspace{-2pt} \sum_{\theta_j \in \Theta_j} \hspace{-2pt} \frac{1 - \pi_j(H^\tau,\hat{a}_j^\tau,\theta_j)}{1 - |A_j|^{-1}}
	\end{equation}
	where $\hat{a}_j^\tau \in \arg\max_{a_j} \pi_j(H^\tau,a_j,\theta_j)$.
\end{_definition}

Both quantities are bounded by 0 and 1. The average overlap describes the similarity of the types, where $\ao_j(H^t) = 0$ means that player $j$'s types (on average) never chose the same action in history $H^t$, whereas $\ao_j(H^t) = 1$ means that they behaved identically. The average stochasticity describes the uncertainty of the types, where $\as_j(H^t) = 0$ means that player $j$'s types (on average) were fully deterministic in the action choices in history $H^t$, whereas $\as_j(H^t) = 1$ means that they chose actions randomly with uniform probability.

We can show that, if the average overlap and stochasticity of player $j$ converge to zero as $t \rightarrow \infty$, then the sum posterior is guaranteed to learn any pure or mixed type distribution:

\begin{_theorem} \label{theo:s-post}
	Let $\Gamma$ be a SBG with a pure or mixed type distribution $\td$. If HBA uses a sum posterior, then, for $t~\rightarrow~\infty$: If $\ao_j(H^t) = 0$ and $\as_j(H^t) = 0$ for all players $j \neq i$, then $\pr (\theta_{-i} | H^t) = \td(\theta_{-i})$ for all $\theta_{-i} \in \Theta_{-i}$.
\end{_theorem}

\begin{proof}
	Throughout this proof, let $t \rightarrow \infty$. The sum posterior is defined as \eqref{eq:post} where $L$ is defined as \eqref{eq:s-post}. Given the definition of $L$, both the numerator and the denominator~in \eqref{eq:post} may be infinite. We invoke L'H\^{o}pital's rule which states that, in such cases, the quotient $\vspace{-3pt}\frac{u(t)}{v(t)}$ is equal to the quotient $\frac{u'(t)}{v'(t)}$ of the respective derivatives with respect to $t$. The derivative of $L$ with respect to $t$ is the average growth per time step, which in general may depend on the history $H^t$ of states and actions. The average growth of $L$ is
	\begin{equation} \label{eq:deriv}
		L'(H^t | \theta_j) = \sum_{a_j \in A_j} F(a_j | H^t) \, \pi_j(H^t,a_j,\theta_j)
	\end{equation}
	where
	\begin{equation}
		F(a_j | H^t) = \sum_{\theta_j \in \Theta_j} \td(\theta_j) \, \pi_j(H^t,a_j,\theta_j)
	\end{equation}
	is the probability of action $a_j$ after history $H^t$, with $\td(\theta_j)$ being the marginal probability that player $j$ is assigned type $\theta_j$. As we will see shortly, we can make an asymptotic growth prediction irrespective of $H^t$. Given that $\ao_j(H^t) = 0$, we can infer that whenever $\pi_j(H^t,a_j,\theta_j) > 0$ for action $a_j$ and type $\theta_j$, then $\pi_j(H^t,a_j,\theta'_j) = 0$ for all other types $\theta'_j \neq \theta_j$. Therefore, we can write \eqref{eq:deriv} as
	\begin{equation}
		L'(H^t | \theta_j) = \td(\theta_j) \sum_{a_j \in A_j} \pi_j(H^t,a_j,\theta_j)^2
	\end{equation}
	Next, given that $\as_j(H^t) = 0$, we know that there exists an action $a_j$ such that $\pi_j(H^t,a_j,\theta_j) = 1$, and therefore we can conclude that $L'(H^t | \theta_j) = \td(\theta_j)$. This shows that the history $H^t$ is irrelevant to the asymptotic growth rate of $L$. Finally, since $\sum_{\theta_j \in \Theta_j} \td(\theta_j) = 1$, we know that the denominator in \eqref{eq:post} will be 1, and we can ultimately conclude that $\pr_j (\theta_j | H^t) = \td(\theta_j)$.
\end{proof}

Theorem~\ref{theo:s-post} explains why the sum posterior converges to the correct type distribution in Example~\ref{ex:product-mixed}. Since the types $\theta_A$ and $\theta_B$ always choose different actions and are completely deterministic (i.e. the average overlap and stochasticity are always zero), the sum posterior is guaranteed to converge to the type distribution. On the other hand, in Example~\ref{ex:sum-pure} the types $\theta_A$ and $\theta_{AB}$ produce an overlap whenever action A is chosen, and $\theta_{AB}$ is completely random. Therefore, the average overlap and stochasticity are always positive, and an incorrect type distribution was learned.

\begin{figure}[t]
	\centering
	\includegraphics[scale=0.53]{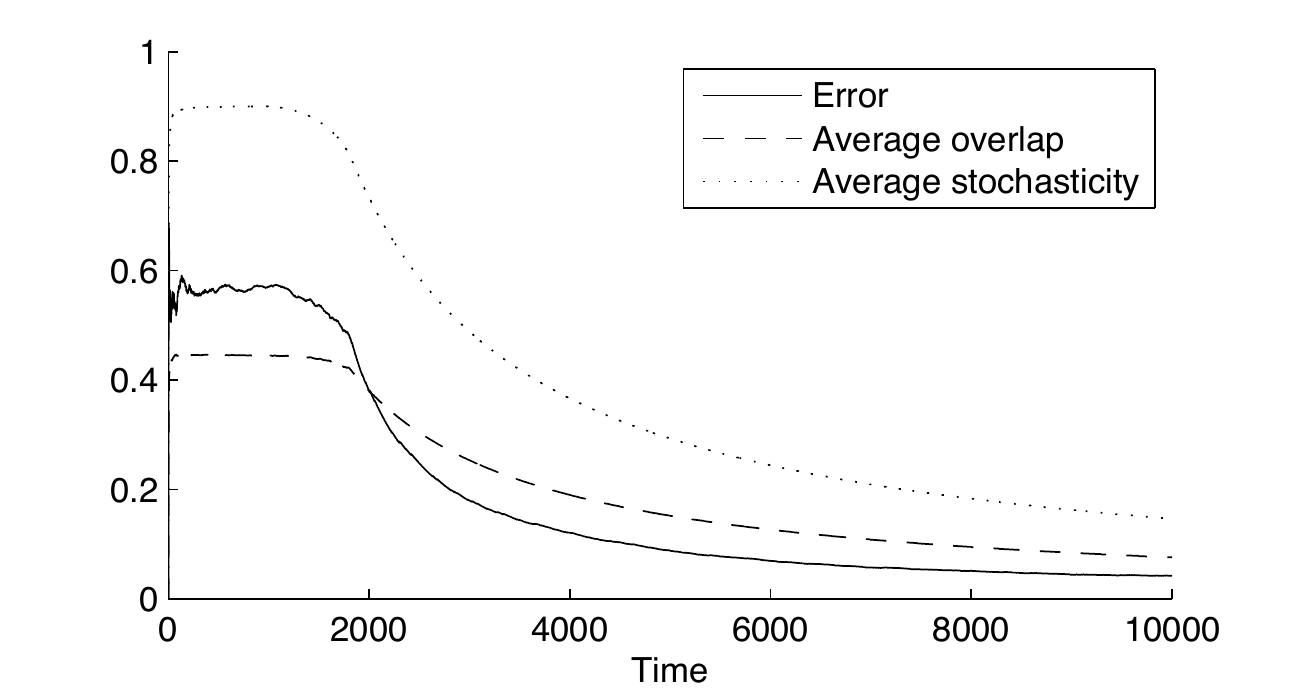}
	\caption{\small Example run in random SBG with 2 players, 10 actions, and 100 states. Player $j$ has 3 reinforcement learning types with $\epsilon$-greedy action selection (decreasing linearly from $\epsilon=0.7$ at $t=1000$, to $\epsilon=0$ at $t = 2000$). The error at time $t$ is $\sum_{\theta_j} | \textnormal{\pr}_j (\theta_j | H^t) - \td(\theta_j) |$ where $\textnormal{\pr}_j$ is the sum posterior.}
	\label{fig:convergence}
\end{figure}

The assumptions made in Theorem~\ref{theo:s-post}, namely that the average overlap and stochasticity converge to zero, require practical justification. First of all, it is important to note that it is only required that these converge to zero \emph{on average} as $t \rightarrow \infty$. This means that in the beginning there may be arbitrary overlap and stochasticity, as long as these go to zero as the game proceeds. In fact, with respect to stochasticity, this is precisely how the exploration-exploitation dilemma \citep{sb1998} is solved in practice: In the early stages, the agent randomises deliberately over its actions in order to obtain more information about the environment (\emph{exploration}) while, as the game proceeds, the agent becomes gradually more deterministic in its action choices so as to maximise its payoffs (\emph{exploitation}). Typical mechanisms which implement this are $\epsilon$-greedy and Softmax/Boltzmann exploration \citep{sb1998}. Figure~\ref{fig:convergence} demonstrates this in a SBG in which player $j$ has 3 reinforcement learning types. The payoffs for the types were such that the average overlap would eventually go to zero.

Regarding the average overlap converging to zero, we believe that this is a property which should be guaranteed \emph{by design}, for the following reason: If the user-defined type space $\Theta_j^*$ is such that there is a constantly high average overlap, then this means that the types $\theta_j^* \in \Theta_j^*$ are in effect very similar. However, types which are very similar are likely to produce very similar trajectories in the planning step of HBA (cf. $\hat{H}$ in \eqref{eq:bne}) and, therefore, constitute redundancy in both time and space. Therefore, we believe it is advisable to use type spaces which have low average overlap.

		\subsection{CORRELATED POSTERIOR}

An implicit assumption in the definition of \eqref{eq:post} is that the type distribution $\td$ can be represented as a product of $n$ independent factors (one for each player), so that $\td(\theta) = \prod_j \td_j(\theta_j)$. Therefore, since the sum posterior is in the form of \eqref{eq:post}, it is in fact only guaranteed to learn \emph{independent} type distributions. This is opposed to \emph{correlated} type distributions, which cannot be represented as a product of $n$ independent factors. Correlated type distributions can be used to specify constraints on type combinations, such as ``player $j$ can only have type $\theta_j$ if player $k$ has type $\theta_k$''. The following example demonstrates how the sum posterior fails to converge to a correlated type distribution:

\begin{_example} \label{ex:correlated}
	Consider a SBG with 3 players. Player 1 is controlled by HBA using a sum posterior. Players 2 and 3 each have two types, $\theta_A$ and $\theta_B$, which are defined as in Example~\ref{ex:product-mixed}. The type distribution $\td$ chooses types with probabilities $\td(\theta_A,\theta_B) = \td(\theta_B,\theta_A) = 0.5$ and $\td(\theta_A,\theta_A) = \td(\theta_B,\theta_B) = 0$. In other words, player 2 can never have the same type as player 3. From the perspective of HBA, each type (and hence action) is chosen with equal probability for both players. Thus, despite the fact that there is zero overlap and stochasticity, the sum posterior will eventually assign probability 0.25 to all constellations of types, which is incorrect. This means that HBA fails to recognise that the other players never choose the same action.
\end{_example}

In this section, we propose a new posterior which can learn any correlated type distribution:

\begin{_definition}
	The \emph{correlated posterior} is defined as
	\begin{equation} \label{eq:c-post}
		\pr (\theta_{-i} | H^t) = \eta \, P(\theta_{-i}) \sum_{\tau = 0}^{t-1} \, \prod_{\theta_j \in \theta_{-i}} \hspace{-3pt} \pi_j(H^\tau,a_j^\tau,\theta_j)
	\end{equation}
	where $P$ specifies prior probabilities (or beliefs) over $\Theta_{-i}$ (analogous to $P_j$) and $\eta$ is a normalisation constant.
\end{_definition}

The correlated posterior is closely related to the sum posterior. In fact, in converges to the true type distribution under the same conditions as the sum posterior:

\begin{_theorem} \label{theo:c-post}
	Let $\Gamma$ be a SBG with a \emph{correlated} type distribution $\td$. If HBA uses the correlated posterior, then, for $t \rightarrow \infty$: If $\ao_j(H^t) = 0$ and $\as_j(H^t) = 0$ for all players $j \neq i$, then $\pr (\theta_{-i} | H^t) = \td(\theta_{-i})$ for all $\theta_{-i} \in \Theta_{-i}$.
\end{_theorem}

\begin{proof}
	Proof is analogous to proof of Theorem~\ref{theo:s-post}.
\end{proof}

It is easy to see that the correlated posterior would learn the correct type distribution in Example~\ref{ex:correlated}. Note that, since it is guaranteed to learn any correlated type distribution, it is also guaranteed to learn any independent type distribution. Therefore, the correlated posterior would also learn the correct type distribution in Example~\ref{ex:product-mixed}. This means that the correlated posterior is \emph{complete} in the sense that it covers the entire spectrum of pure/mixed and independent/correlated type distributions. However, this completeness comes at a higher computational complexity. While the sum posterior is in $O(n \max_j |\Theta_j|)$ time and space, the correlated posterior is in $O(\max_j |\Theta_j|^n)$ time and space. In practice, however, the time complexity can be reduced drastically by computing the probabilities $\pi_j(H^\tau,a_j^\tau,\theta_j)$ only once for each $j$ and $\theta_j \in \Theta_j$ (as in the sum posterior), and then reusing them in subsequent computations.

	\section{INACCURATE TYPE SPACES} \label{sec:type-space}

Each user-defined type $\theta^*_j$ in $\Theta^*_j$ is a hypothesis by the user regarding how player $j$ might behave. Therefore, $\Theta^*_j$ may be \emph{inaccurate} in the sense that none of the types therein accurately predict the observed behaviour of player $j$. This is demonstrated in the following example:

\begin{_example} \label{ex:inaccurate}
	Consider a SBG with two players and actions L and R. Player 1 is controlled by HBA while player 2 has a single type, $\theta_{LR}$, which chooses L,R,L,R, etc. HBA is provided with $\Theta_j^* = \left\{ \theta^*_R , \theta^*_{LRR} \right\}$, where $\theta^*_R$ always chooses R while $\theta^*_{LRR}$ chooses L,R,R,L,R,R etc. Both user-defined types are inaccurate in the sense that they predict player 2's actions in only $\approx$ 50\% of the game.
\end{_example}

Two important theoretical questions in this context are how closely the user-defined type spaces $\Theta^*_j$ have to approximate the real type spaces $\Theta_j$ in order for HBA to be able to \textit{(1)} solve the task (i.e. bring the SBG into a terminal state), and \textit{(2)} achieve maximum payoffs. These questions are closely related to the notions of \emph{flexibility} and \emph{efficiency} \citep{ar2013} which, respectively, correspond to the probability of termination and the average payoff per time step. In this section, we are primarily concerned with question 1, and we are concerned with question 2 only in so far as that we want to solve the task in minimal time. (Since reducing the time until termination will increase the average payoff per time step, i.e. increase efficiency.) This focus is formally captured by the following assumption, which we make throughout this section:

\begin{_assumption} \label{ass:terminal-payoffs}
	Let player $i$ be controlled by HBA, then $u_i(s,a,\theta_i) = 1 \text{ iff. } s \in \bar{S} \text{, else } 0$.
\end{_assumption}

Assumption~\ref{ass:terminal-payoffs} specifies that we are only interested in reaching a terminal state, since this is the only way to obtain a none-zero payoff. In our analysis, we consider discount factors $\gamma$ (cf. Algorithm~\ref{alg:hba}) with $\gamma = 1$ and $\gamma < 1$. While all our results hold for both cases, there is an important distinction: If $\gamma = 1$, then the expected payoffs \eqref{eq:bne} correspond to the actual probability that the following state can lead to (or is) a terminal state (we call this the \emph{success rate}), whereas this is not necessarily the case if $\gamma < 1$. This is since $\gamma < 1$ tends to prefer shorter paths, which means that actions with lower success rates may be preferred if they lead to faster termination. Therefore, if $\gamma = 1$ then HBA is solely interested in termination, and if $\gamma < 1$ then it is interested in \emph{fast} termination, where lower $\gamma$ prefers faster termination.

		\subsection{METHODOLOGY OF ANALYSIS}

Given a SBG $\Gamma$, we define the \emph{ideal process}, $X$, as the process induced by $\Gamma$ in which player $i$ is controlled by HBA and in which HBA always knows the current and all future types of all players. Then, given a posterior $\pr$ and user-defined type spaces $\Theta^*_j$ for all $j \neq i$, we define the \emph{user process}, $Y$, as the process induced by $\Gamma$ in which player $i$ is controlled by HBA (same as in $X$) and in which HBA uses $\pr$ and $\Theta^*_j$ in the usual way. Thus, the only difference between $X$ and $Y$ is that $X$ can always predict the player types whereas $Y$ approximates this knowledge through $\pr$ and $\Theta^*_j$. We write $E_{s^t}^{a_i}(H^t | C)$ to denote the expected payoff (as defined by \eqref{eq:bne}) of action $a_i$ in state $s^t$ after history $H^t$, in process $C \in \left\{ X,Y \right\}$.

The idea is that $X$ constitutes the ideal solution in the sense that $E_{s^t}^{a_i}(H^t | X)$ corresponds to the \emph{actual} expected payoff, which means that HBA chooses the truly best-possible actions in $X$. This is opposed to $E_{s^t}^{a_i}(H^t | Y)$, which is merely the \emph{estimated} expected payoff based on $\pr$ and $\Theta^*_j$, so that HBA may choose suboptimal actions in $Y$. The methodology of our analysis is to specify what relation $Y$ must have to $X$ to satisfy certain guarantees for termination.

We specify such guarantees in PCTL \citep{hj1994}, a probabilistic modal logic which also allows for the specification of time constraints. PCTL expressions are interpreted over infinite histories in labelled transition systems with atomic propositions (i.e. Kripke structures). In order to interpret PCTL expressions over $X$ and $Y$, we make the following modifications without loss of generality: Firstly, any terminal state $\bar{s} \in \bar{S}$ is an \emph{absorbing} state, meaning that if a process is in $\bar{s}$, then the next state will be $\bar{s}$ with probability 1 and all players receive a zero payoff. Secondly, we introduce the atomic proposition \texttt{term} and label each terminal state with it, so that \texttt{term} is true in $s$ if and only if $s \in \bar{S}$.

We will use the following two PCTL expressions:
\begin{equation*}
	F^{\leq t}_{\succ p} \texttt{term}, \ F^{< \infty}_{\succ p} \texttt{term}
\end{equation*}
where $t \in \mathbb{N}$, $p \in [0,1]$, and $\succ \in \left\{ >, \geq \right\}$.

$F^{\leq t}_{\succ p} \texttt{term}$ specifies that, given a state $s$, with a probability of $\succ p$ a state $s'$ will be reached from $s$ within $t$ time steps such that $s'$ satisfies $\texttt{term}$. The semantics of $F^{< \infty}_{\succ p} \texttt{term}$ are similar except that $s'$ will be reached in arbitrary but finite time. We write $s \models_C \phi$ to say that a state $s$ satisfies the PCTL expression $\phi$ in process $C \in \left\{ X,Y \right\}$.

		\subsection{CRITICAL TYPE SPACES}

In the following section, we sometimes assume that the user-defined type spaces $\Theta^*_j$ are \emph{uncritical}:

\begin{_definition}
	The user-defined type spaces $\Theta^*_j$ are \emph{critical} if there is $S^c \subseteq S \setminus \bar{S}$ which satisfies:
	\vspace{-6pt}
	\begin{enumerate}[leftmargin=\leftm,itemsep=0pt]
		\item For each $H^t \in \mathbb{H}$ with $s^t \in S^c$, there is $a_i \in A_i$ such that $E_{s^t}^{a_i}(H^t | Y) > 0$ and $E_{s^t}^{a_i}(H^t | X) > 0$
		\item There is a positive probability that $Y$ may eventually get into a state $s^c \in S^c$ from the initial state $s^0$
		\item If Y is in a state in $S^c$, then with probability 1 it will always be in a state in $S^c$ (i.e. it will never leave $S^c$)
	\end{enumerate}
	\vspace{-5pt}
	We say $\Theta^*_j$ are \emph{uncritical} if they are not critical.
\end{_definition}

Intuitively, critical type spaces have the potential to lead HBA into a state space in which it \emph{believes} it chooses the right actions to solve the task, while other actions are \emph{actually} required to solve the task. The only effect that its actions have is to induce an infinite cycle, due to a critical inconsistency between the user-defined and true type spaces. The following example demonstrates this:

\begin{_example} \label{ex:critical}
	Recall Example~\ref{ex:inaccurate} and let the task be to choose the same action as player $j$. Then, $\Theta_j^*$ is uncritical because HBA will always solve the task at $t = 1$, regardless of the posterior and despite the fact that $\Theta_j^*$ is inaccurate. Now, assume that $\Theta_j^* = \left\{ \theta^*_{RL} \right\}$ where $\theta^*_{RL}$ chooses actions R,L,R,L etc. Then, $\Theta_j^*$ is critical since HBA will always choose the opposite action of player $j$, thinking that it would solve the task, when a different action would actually solve it.
\end{_example}

A practical way to ensure that the type spaces $\Theta^*_j$ are (eventually) uncritical is to include methods for opponent modelling in each $\Theta^*_j$ (e.g. as in \citep{ar2013,bsk2011}). If the opponent models are guaranteed to learn the correct behaviours, then the type spaces $\Theta^*_j$ are guaranteed to become uncritical. In Example~\ref{ex:critical}, any standard modelling method would eventually learn that the true strategy of player $j$ is $\theta_{LR}$. As the model becomes more accurate, the posterior gradually shifts towards it and eventually allows HBA to take the right action.

		\subsection{TERMINATION GUARANTEES}

Our first guarantee states that if $X$ has a positive probability of solving the task, then so does $Y$:

\begin{_property} \label{prop1}
	$s^0 \models_X F^{< \infty}_{> 0} \texttt{term} \ \Rightarrow\ s^0 \models_Y F^{< \infty}_{> 0} \texttt{term}$
\end{_property}

We can show that Property~\ref{prop1} holds if the user-defined type spaces $\Theta^*_j$ are uncritical and if $Y$ only chooses actions for player $i$ with positive expected payoff in $X$.

Let $\mathbb{A}(H^t|C)$ denote the set of actions that process $C$ may choose from in state $s^t$ after history $H^t$, i.e. $\mathbb{A}(H^t|C) = \arg\max_{a_i} E_{s^t}^{a_i}(H^t | C)$ (cf. step 3 in Algorithm~\ref{alg:hba}).

\begin{_theorem} \label{theo:prop1}
	Property \ref{prop1} holds if $\Theta^*_j$ are uncritical and
	\begin{equation} \label{eq:prop1}
		\forall H^t \hspace{-2pt} \in \mathbb{H} \ \forall a_i \in \mathbb{A}(H^t | Y) : E_{s^t}^{a_i}(H^t | X) > 0
	\end{equation}
\end{_theorem}

\begin{proof}
	Assume $s^0 \models_X F^{< \infty}_{> 0} \texttt{term}$. Then, we know that $X$ chooses actions $a_i$ which \emph{may} lead into a state $s'$ such that $s' \models_X F^{< \infty}_{> 0} \texttt{term}$, and the same holds for all such states $s'$. Now, given \eqref{eq:prop1} it is tempting to infer the same result for $Y$, since $Y$ only chooses actions $a_i$ which have positive expected payoff in $X$ and, therefore, could truly lead into a terminal state. However, \eqref{eq:prop1} alone is not sufficient to infer $s' \models_Y F^{< \infty}_{> 0} \texttt{term}$ because of the special case in which $Y$ chooses actions $a_i$ such that $E_{s^t}^{a_i}(H^t | X) > 0$ but without ever reaching a terminal state. This is why we require that the user-defined type spaces $\Theta^*_j$ are uncritical, which prevents this special case. Thus, we can infer that $s' \models_Y F^{< \infty}_{> 0} \texttt{term}$, and hence Property~\ref{prop1} holds.
\end{proof}

The second guarantee states that if $X$ always solves the task, then so does $Y$:

\begin{_property} \label{prop2}
	$s^0 \models_X F^{< \infty}_{\geq 1} \texttt{term} \ \Rightarrow\ s^0 \models_Y F^{< \infty}_{\geq 1} \texttt{term}$
\end{_property}

We can show that Property~\ref{prop2} holds if the user-defined type spaces $\Theta^*_j$ are uncritical and if $Y$ only chooses actions for player $i$ which lead to states into which $X$ may get as well.

Let $\mu(H^t,s|C)$ be the probability that process $C$ transitions into state $s$ from state $s^t$ after history $H^t$, i.e. $\mu(H^t,s|C) = \frac{1}{|\mathbb{A}|} \hspace{-2pt} \sum_{a_i \in \mathbb{A}} \sum_{a_{-i}} T(s^t,\langle a_i,a_{-i} \rangle,s) \prod_j \pi_j(H^t,a_j,\theta_j^t)$ with $\mathbb{A} \equiv \mathbb{A}(H^t|C)$, and let $\mu(H^t,S'|C) = \sum_{s \in S'} \mu(H^t,s|C)$ for $S' \subset S$.

\begin{_theorem} \label{theo:prop2}
	Property \ref{prop2} holds if $\Theta^*_j$ are uncritical and
	\begin{equation} \label{eq:prop2}
		\forall H^t \hspace{-2pt} \in \mathbb{H} \ \forall s \in S : \mu(H^t,s|Y) > 0 \Rightarrow \mu(H^t,s|X) > 0
	\end{equation}
\end{_theorem}

\begin{proof}
	The fact that $s^0 \models_X F^{< \infty}_{\geq 1} \texttt{term}$ means that, throughout the process, $X$ only transitions into states $s$ with $s \models_X F^{< \infty}_{\geq 1} \texttt{term}$. As before, it is tempting to infer the same result for $Y$ based on \eqref{eq:prop2}, since it only transitions into states which have maximum success rate in $X$. However, \eqref{eq:prop2} alone is not sufficient since $Y$ may choose actions such that \eqref{eq:prop2} holds true but $Y$ will never reach a terminal state. Nevertheless, since the user-defined type spaces $\Theta_j^*$ are uncritical, we know that this special case will not occur, and hence Property~\ref{prop2} holds.
\end{proof}

We note that, in both Properties \ref{prop1} and \ref{prop2}, the reverse direction holds true regardless of Theorems \ref{theo:prop1} and \ref{theo:prop2}. Furthermore, we can combine the requirements of Theorems \ref{theo:prop1} and \ref{theo:prop2} to ensure that both properties hold.

The next guarantee subsumes the previous guarantees by stating that $X$ and $Y$ have the same minimum probability of solving the task:

\begin{_property} \label{prop3}
	$s^0 \models_X F^{< \infty}_{\geq p} \texttt{term} \ \Rightarrow\ s^0 \models_Y F^{< \infty}_{\geq p} \texttt{term}$
\end{_property}

We can show that Property~\ref{prop3} holds if the user-defined type spaces $\Theta^*_j$ are uncritical and if $Y$ only chooses actions for player $i$ which $X$ might have chosen as well.

Let $R(a_i,H^t|C)$ be the \emph{success rate} of action $a_i$, formally $R(a_i,H^t|C) = E_{s^t}^{a_i}(H^t | C)$ with $\gamma = 1$ (so that it corresponds to the actual \emph{probability} with which $a_i$ may lead to termination in the future). Define $X_{\min}$ and $X_{\max}$ to be the processes which for each $H^t$ choose actions $a_i \in \mathbb{A}(H^t|X)$ with, respectively, minimal and maximal success rate $R(a_i,H^t|X)$.

\begin{_theorem} \label{theo:prop3}
	If $\Theta^*_j$ are uncritical and
	\begin{equation} \label{eq:prop3}
		\forall H^t \hspace{-2pt} \in \mathbb{H} : \mathbb{A}(H^t|Y) \subseteq \mathbb{A}(H^t|X)
	\end{equation}

	\vspace{-13pt}

	then

	(i) for $\gamma = 1$: Proposition~\ref{prop3} holds in both directions

	(ii) for $\gamma < 1$: $s^0 \models_X F^{< \infty}_{\geq p} \texttt{term} \ \Rightarrow\ s^0 \models_Y F^{< \infty}_{\geq p'} \texttt{term}$

	with $p_{\min} \leq q \leq p_{\max}$ for $q \in \left\{ p,p' \right\}$, where $p_{\min}$ and $p_{\max}$ are the highest probabilities such that $s^0 \models_{X_{\min}} F^{< \infty}_{\geq p_{\min}} \texttt{term}$ and $s^0 \models_{X_{\max}} F^{< \infty}_{\geq p_{\max}} \texttt{term}$.
\end{_theorem}

\begin{proof}
	(i): Since $\gamma = 1$, all actions $a_i \in \mathbb{A}(H^t|X)$ have the same success rate for a given $H^t$, and given \eqref{eq:prop3} we know that $Y$'s actions always have the same success rate as $X$'s actions. Provided that the type spaces $\Theta_j^*$ are uncritical, we can conclude that Property~\ref{prop3} must hold, and for the same reasons the reverse direction must hold as well.
	
	(ii): Since $\gamma < 1$, the actions $a_i \in \mathbb{A}(H^t|X)$ may have different success rates. The lowest and highest chances that $X$ solves the task are precisely modelled by $X_{\min}$ and $X_{\max}$, and given \eqref{eq:prop3} and the fact that $\Theta^*_j$ are uncritical, the same holds for $Y$. Therefore, we can infer the common bound $p_{\min} \leq \left\{ p,p' \right\} \leq p_{\max}$ as defined in Theorem~\ref{theo:prop3}.
\end{proof}

Properties \ref{prop1} to \ref{prop3} are \emph{indefinite} in the sense that they make no restrictions on time requirements. Our fourth and final guarantee subsumes all previous guarantees and states that if there is a probability $p$ such that $X$ terminates \emph{within} $t$ time steps, then so does $Y$ for the same $p$ and $t$:

\begin{_property} \label{prop4}
	$s^0 \models_X F^{\leq t}_{\geq p} \texttt{term} \ \Rightarrow\ s^0 \models_Y F^{\leq t}_{\geq p} \texttt{term}$
\end{_property}

We believe that Property~\ref{prop4} is an adequate criterion of optimality for the type spaces $\Theta_j^*$ since, if it holds, $\Theta_j^*$ must approximate $\Theta_j$ in a way which allows HBA to plan (almost) as accurately --- in terms of solving the task --- as the ``ideal'' HBA in $X$ which always knows the true types.

What relation must $Y$ have to $X$ to satisfy Property~\ref{prop4}? The fact that $Y$ and $X$ are processes over state transition systems means we can draw on methods from the model checking literature to answer this question. Specifically, we will use the concept of \emph{probabilistic bisimulation} \citep{ls1991}, which we here define in the context of our work:

\begin{_definition}
	A \emph{probabilistic bisimulation} between $X$ and $Y$ is an equivalence relation $B \subseteq S \times S$ such that

	(i) $(s^0,s^0) \in B$

	(ii) $s_X \models_X \texttt{term} \Leftrightarrow s_Y \models_Y \texttt{term}$ for all $(s_X,s_Y) \in B$

	(iii) $\mu(H_X^t,\hat{S}|X) \hspace{-1pt}=\hspace{-1pt} \mu(H_Y^t,\hat{S}|Y)$ for any histories $H_X^t,H_Y^t$ with $(s_X^t,s_Y^t) \in B$ and all equivalence classes $\hat{S}$ under $B$.
\end{_definition}

Intuitively, a probabilistic bisimulation states that $X$ and $Y$ do (on average) match each other's transitions. Our definition of probabilistic bisimulation is most general in that it does not require that transitions are matched by the same action or that related states satisfy the same atomic propositions other than termination. However, we do note that other definitions exist that make such additional requirements, and our results hold for each of these refinements.

The main contribution of this section is to show that the optimality criterion expressed by Property~\ref{prop4} holds in \emph{both directions} if there is a probabilistic bisimulation between $X$ and $Y$. Thus, we offer an alternative formal characterisation of optimality for the user-defined type spaces $\Theta_j^*$:

\vspace{5pt}

\begin{_theorem} \label{theo:prop4}
	Property~\ref{prop4} holds in both directions if there is a probabilistic bisimulation between $X$ and $Y$.
\end{_theorem}

\begin{proof}
	First of all, we note that, strictly speaking, the standard definitions of bisimulation (e.g. \citep{b1996,ls1991}) assume the Markov property, which means that the next state of a process depends only on the current state of the process. In contrast, we consider the more general case in which the next state may depend on the history $H^t$ of previous states and joint actions (since the player strategies $\pi_j$ depend on $H^t$). However, one can always enforce the Markov property \emph{by design}, i.e. by augmenting the state space $S$ to account for the relevant factors of the past. In fact, we could postulate that the histories as a whole constitute the states of the system, i.e. $S = \mathbb{H}$. Therefore, to simplify the exposition, we assume the Markov property and we write $\mu(s,\hat{S}|C)$ to denote the cumulative probability that $C$ transitions from state $s$ into any state in $\hat{S}$.

Given the Markov property, the fact that $B$ is an equivalence relation, and $\mu(s_X,\hat{S}|X) = \mu(s_Y,\hat{S}|Y)$ for $(s_X,s_Y) \in B$, we can represent the dynamics of $X$ and $Y$ in a common graph, such as the following one:

	\begin{center}
		\begin{tikzpicture}[every node/.style={inner sep=0pt,minimum size=20pt}]
			\small
			\node (n0) [style={circle,draw},fill=black!15,label=left:$s^0 \in\ $] {$\hat{S}_0$};
			\node (n1) [style={circle,draw},right=of n0] {$\hat{S}_1$};
			\node (n2) [style={circle,draw},right=of n0,below=of n1] {$\hat{S}_2$};
			\node (n3) [style={circle,draw},right=of n1] {$\hat{S}_3$};
			\node (n4) [style={circle,draw},right=of n2,below=of n3] {$\hat{S}_4$};
			\node (n5) [style={circle,draw},right=of n3] {$\hat{S}_5$};
			\node (n6) [style={circle,draw},right=of n4,fill=black!15,label=right:$\ \equiv \bar{S}$] {$\hat{S}_6$};
			\draw (n0) -- (n1) [->] node [style={minimum size=10pt},midway,above] {$\mu_{01}$};
			\draw (n0) -- (n2) [->] node [midway,left] {$\mu_{02}$} node [swap] [midway,right] {$\hspace{22pt}\mu_{14}$};
			\draw (n1) -- (n3) [->] node [style={minimum size=10pt},midway,above] {$\mu_{13}$};
			\draw (n2) -- (n4) [->] node [style={minimum size=10pt},midway,below] {$\mu_{24}$};
			\draw (n3) -- (n5) [->] node [style={minimum size=10pt},midway,above] {$\mu_{35}$};
			\draw (n3) -- (n6) [->] node [midway,left] {$\mu_{41}\hspace{21pt}$} node [swap] [midway,right] {$\hspace{3pt}\mu_{36}$};
			\draw (n4) -- (n6) [->] node [style={minimum size=10pt},midway,below] {$\mu_{46}$};
			\draw [->] (n1) to [bend right=15] (n4);
			\draw [->] (n4) to [bend right=15] (n1);
			\draw [->] (n5) to [loop right] (n5);
		\end{tikzpicture}
	\end{center}

	The nodes correspond to the equivalence classes under $B$. A directed edge from $\hat{S}_a$ to $\hat{S}_b$ specifies that there is a positive probability $\mu_{ab} = \mu(s_X,\hat{S}_b|X) = \mu(s_Y,\hat{S}_b|Y)$ that $X$ and $Y$ transition from states $s_X,s_Y \in \hat{S}_a$ to states $s'_X,s'_Y \in \hat{S}_b$. Note that $s_X,s_Y$ and $s'_X,s'_Y$ need not be equal but merely equivalent, i.e. $(s_X,s_Y) \in B$ and $(s'_X,s'_Y) \in B$. There is one node ($\hat{S}_0$) that contains the initial state $s^0$ and one node ($\hat{S}_6$) that contains all terminal states $\bar{S}$ and no other states. This is because once $X$ and $Y$ reach a terminal state they will always stay in it (i.e. $\mu(s,\bar{S}|X) = \mu(s,\bar{S}|Y) = 1$ for $s \in \bar{S}$) and since they are the only states that satisfy \texttt{term}. Thus, the graph starts in $\hat{S}_0$ and terminates (if at all) in $\hat{S}_6$.

	Since the graph represents the dynamics of both $X$ and $Y$, it is easy to see that Property~\ref{prop4} must hold in both directions. In particular, the probabilities that $X$ and $Y$ are in node $\hat{S}$ at time $t$ are identical. One simply needs to add the probabilities of all directed paths of length $t$ which end in $\hat{S}$ (provided that such paths exist), where the probability of a path is the product of the $\mu_{ab}$ along the path. Therefore, $X$ and $Y$ terminate with equal probability, and on average within the same number of time steps.
\end{proof}

Some remarks to clarify the usefulness of this result: First of all, in contrast to Theorems~\ref{theo:prop1} to \ref{theo:prop3}, Theorem~\ref{theo:prop4} does not explicitly require $\Theta_j^*$ to be uncritical. In fact, this is implicit in the definition of probabilistic bisimulation. Moreover, while the other theorems relate $Y$ and $X$ for identical histories $H^t$, Theorem~\ref{theo:prop4} relates $Y$ and $X$ for \emph{related} histories $H_Y^t$ and $H_X^t$, making it more generally applicable. Finally, Theorem~\ref{theo:prop4} has an important practical implication: it tells us that we can use efficient methods for model checking (e.g. \citep{b1996,ls1991}) to verify optimality of $\Theta_j^*$. In fact, it can be shown that for Property~\ref{prop4} to hold (albeit not in the other direction) it suffices that $Y$ be a \emph{probabilistic simulation} \citep{b1996} of $X$, which is a coarser preorder than probabilistic bisimulation. However, algorithms for checking probabilistic simulation (e.g. \citep{b1996}) are computationally much more expensive (and fewer) than those for probabilistic bisimulation, hence their practical use is currently limited.

	\section{CONCLUSION} \label{sec:conc}


This paper complements works such as [Albrecht and Rama- moorthy, 2013, Barrett et al., 2011, Carmel and Markovitch, 1999] --- with a focus on HBA due to its generality --- by providing answers to two important theoretical questions: ``Under what conditions does HBA learn the type distribution of the game?'' and ``How accurate must the user-defined type spaces be for HBA to solve its task?'' With respect to the first question, we analyse the convergence properties of two existing posteriors and propose a new posterior which can learn correlated type distributions. This provides the user with formal reasons as to which posterior should be chosen for the problem at hand. With respect to the second question, we describe a methodology in which we analyse the requirements of several termination guarantees, and we provide a novel characterisation of optimality which is based on the notion of probabilistic bisimulation. This gives the user a formal yet practically useful criterion of what constitutes optimal type spaces. The results of this work improve our understanding of how a method such as HBA can be used to effectively solve agent interaction problems in which the behaviour of other agents is not a priori known.

There are several interesting directions for future work. For instance, it is unclear what effect the prior probabilities $P_j$ have on the performance of HBA, and if a criterion for optimal $P_j$ could be derived. Furthermore, since our convergence proofs in Section~\ref{sec:td} are asymptotic, it would be interesting to know if useful finite-time error bounds \text{exist}. Finally, our analysis in Section~\ref{sec:type-space} is general in the sense that it applies to any posterior. This could be refined by an analysis which commits to a specific posterior.

	\bibliographystyle{plainnat}
	\bibliography{uai14}

\begin{thebibliography}{19}
\providecommand{\natexlab}[1]{#1}
\providecommand{\url}[1]{\texttt{#1}}
\expandafter\ifx\csname urlstyle\endcsname\relax
  \providecommand{\doi}[1]{doi: #1}\else
  \providecommand{\doi}{doi: \begingroup \urlstyle{rm}\Url}\fi

\bibitem[Albrecht and Ramamoorthy(2012)]{ar2012}
S.~Albrecht and S.~Ramamoorthy.
\newblock Comparative evaluation of {MAL} algorithms in a diverse set of ad hoc
  team problems.
\newblock In \emph{Proceedings of the 11th International Conference on
  Autonomous Agents and Multiagent Systems}, 2012.

\bibitem[Albrecht and Ramamoorthy(2013)]{ar2013}
S.~Albrecht and S.~Ramamoorthy.
\newblock A game-theoretic model and best-response learning method for ad hoc
  coordination in multiagent systems.
\newblock In \emph{Proceedings of the 12th International Conference on
  Autonomous Agents and Multiagent Systems}, 2013.

\bibitem[Albrecht and Ramamoorthy(2014)]{ar2014app}
S.~Albrecht and S.~Ramamoorthy.
\newblock On convergence and optimality of best-response learning with policy
  types in multiagent systems -- {A}ppendix, 2014.
\newblock
  \newline\textit{http://rad.inf.ed.ac.uk/data/publications/2014/uai14app.pdf}.

\bibitem[Avrahami-Zilberbrand and Kaminka(2007)]{ak2007}
D.~Avrahami-Zilberbrand and G.~Kaminka.
\newblock Incorporating observer biases in keyhole plan recognition
  (efficiently!).
\newblock In \emph{Proceedings of the 22nd AAAI Conference on Artificial
  Intelligence}, 2007.

\bibitem[Baier(1996)]{b1996}
C.~Baier.
\newblock Polynomial time algorithms for testing probabilistic bisimulation and
  simulation.
\newblock In \emph{Proceedings of the 8th International Conference on Computer
  Aided Verification, Lecture Notes in Computer Science}, volume 1102, pages
  38--49. Springer, 1996.

\bibitem[Barrett et~al.(2011)Barrett, Stone, and Kraus]{bsk2011}
S.~Barrett, P.~Stone, and S.~Kraus.
\newblock Empirical evaluation of ad hoc teamwork in the pursuit domain.
\newblock In \emph{Proceedings of the 10th International Conference on
  Autonomous Agents and Multiagent Systems}, 2011.

\bibitem[Bonchek-Dokow et~al.(2009)Bonchek-Dokow, Kaminka, and
  Domshlak]{bkd2009}
E.~Bonchek-Dokow, G.~Kaminka, and C.~Domshlak.
\newblock Distinguishing between intentional and unintentional sequences of
  actions.
\newblock In \emph{Proceedings of the 9th International Conference on Cognitive
  Modeling}, 2009.

\bibitem[Bowling and McCracken(2005)]{bm2005}
M.~Bowling and P.~McCracken.
\newblock Coordination and adaptation in impromptu teams.
\newblock In \emph{Proceedings of the 20th National Conference on Artificial
  Intelligence}, 2005.

\bibitem[Carberry(2001)]{c2001}
S.~Carberry.
\newblock Techniques for plan recognition.
\newblock \emph{User Modeling and User-Adapted Interaction}, 11\penalty0
  (1-2):\penalty0 31--48, 2001.

\bibitem[Carmel and Markovitch(1999)]{cm1999}
D.~Carmel and S.~Markovitch.
\newblock Exploration strategies for model-based learning in multi-agent
  systems: Exploration strategies.
\newblock \emph{Autonomous Agents and Multi-Agent Systems}, 2\penalty0
  (2):\penalty0 141--172, 1999.

\bibitem[Doshi et~al.(2009)Doshi, Zeng, and Chen]{dzc2009}
P.~Doshi, Y.~Zeng, and Q.~Chen.
\newblock Graphical models for interactive {POMDPs}: representations and
  solutions.
\newblock \emph{Autonomous Agents and Multi-Agent Systems}, 18\penalty0
  (3):\penalty0 376--416, 2009.

\bibitem[Gilboa and Schmeidler(2001)]{gs2001}
I.~Gilboa and D.~Schmeidler.
\newblock \emph{A theory of case-based decisions}.
\newblock Cambridge University Press, 2001.

\bibitem[Gmytrasiewicz and Doshi(2005)]{gd2005}
P.~Gmytrasiewicz and P.~Doshi.
\newblock A framework for sequential planning in multiagent settings.
\newblock \emph{Journal of Artificial Intelligence Research}, 24\penalty0
  (1):\penalty0 49--79, 2005.

\bibitem[Gmytrasiewicz and Durfee(2000)]{gd2000}
P.~Gmytrasiewicz and E.~Durfee.
\newblock Rational coordination in multi-agent environments.
\newblock \emph{Autonomous Agents and Multi-Agent Systems}, 3\penalty0
  (4):\penalty0 319--350, 2000.

\bibitem[Hansson and Jonsson(1994)]{hj1994}
H.~Hansson and B.~Jonsson.
\newblock A logic for reasoning about time and reliability.
\newblock \emph{Formal Aspects of Computing}, 6\penalty0 (5):\penalty0
  512--535, 1994.

\bibitem[Kalai and Lehrer(1993)]{kl1993}
E.~Kalai and E.~Lehrer.
\newblock Rational learning leads to {N}ash equilibrium.
\newblock \emph{Econometrica}, pages 1019--1045, 1993.

\bibitem[Larsen and Skou(1991)]{ls1991}
K.~Larsen and A.~Skou.
\newblock Bisimulation through probabilistic testing.
\newblock \emph{Information and Computation}, 94\penalty0 (1):\penalty0 1--28,
  1991.

\bibitem[Stone et~al.(2010)Stone, Kaminka, Kraus, and Rosenschein]{skkr2010}
P.~Stone, G.~Kaminka, S.~Kraus, and J.~Rosenschein.
\newblock Ad hoc autonomous agent teams: Collaboration without
  pre-coordination.
\newblock In \emph{Proceedings of the 24th AAAI Conference on Artificial
  Intelligence}, 2010.

\bibitem[Sutton and Barto(1998)]{sb1998}
R.~Sutton and A.~Barto.
\newblock \emph{Reinforcement learning: An introduction}.
\newblock The MIT Press, 1998.

\end{thebibliography}

\end{document}